\def\year{2020}\relax
\documentclass[letterpaper]{article} 
\usepackage{aaai20}  
\usepackage{times}  
\usepackage{helvet} 
\usepackage{courier}  
\usepackage[hyphens]{url}  
\usepackage{graphicx} 
\usepackage{graphicx}  
\usepackage{tabu}
\usepackage{array,booktabs,makecell}
\usepackage{subfig}
\usepackage[reqno]{amsmath}
\usepackage{amssymb}
\usepackage{subfig}
\usepackage{textcomp}
\allowdisplaybreaks[1]

\newcommand{\dashrule}[1][black]{%
  \color{#1}\rule[\dimexpr.5ex-.2pt]{4pt}{.4pt}\xleaders\hbox{\rule{4pt}{0pt}\rule[\dimexpr.5ex-.2pt]{4pt}{.4pt}}\hfill\kern0pt%
}
\newcommand{\newcite}[1]{\citeauthor{#1} \shortcite{#1}}
\title{Syntactically Look-Ahead Attention Network for Sentence Compression}
\author{
    Hidetaka Kamigaito, Manabu Okumura\\
    Institute of Innovative Research, Tokyo Institute of Technology\\
 kamigaito@lr.pi.titech.ac.jp, oku@pi.titech.ac.jp
}

\begin{document}

\maketitle

\begin{abstract}
Sentence compression is the task of compressing a long sentence into a short one by deleting redundant words.
In sequence-to-sequence (Seq2Seq) based models, the decoder unidirectionally decides to retain or delete words.
Thus, it cannot usually explicitly capture the relationships between decoded words and unseen words that will be decoded in the future time steps.
Therefore, to avoid generating ungrammatical sentences, the decoder sometimes drops important words in compressing sentences. 
To solve this problem, we propose a novel Seq2Seq model, \textit{syntactically look-ahead attention network} (SLAHAN), that can generate informative summaries by explicitly tracking both dependency parent and child words during decoding and capturing important words that will be decoded in the future.
The results of the automatic evaluation on the Google sentence compression
dataset showed that SLAHAN achieved the best kept-token-based-F1, ROUGE-1, ROUGE-2 and ROUGE-L scores of 85.5, 79.3, 71.3 and 79.1, respectively.
SLAHAN also improved the summarization performance on longer sentences.
Furthermore, in the human evaluation, SLAHAN improved informativeness without losing readability.
\end{abstract}

\section{Introduction}

Sentence compression is the task of producing a shorter sentence by deleting words in the input sentence while preserving its grammaticality and important information. To compress a sentence so that it is still grammatical, tree trimming methods \cite{jing:2000:ANLP,knight2000statistics,bergkirkpatrick-gillick-klein:2011:ACL-HLT2011,filippova-altun:2013:EMNLP} have been utilized. However, these methods often suffer from parsing errors.
As an alternative, \newcite{filippova-EtAl:2015:EMNLP} proposed a method based on sequence-to-sequence (Seq2Seq) models that do not rely on parse trees but produce fluent compression. However, the vanilla Seq2Seq model has a problem that it is not so good for compressing longer sentences.

To solve the problem, \newcite{kamigaito-etal-2018-higher} expanded Seq2Seq models to capture the relationships between long-distance
words through recursively tracking dependency parents from a word with their recursive attention module.
Their model learns dependency trees and compresses sentences jointly to avoid the effect of parsing errors.
This improvement enables their model to compress a sentence while preserving the important words and its fluency.

\begin{figure*}[t]
\centering
\includegraphics[width=0.9\textwidth]{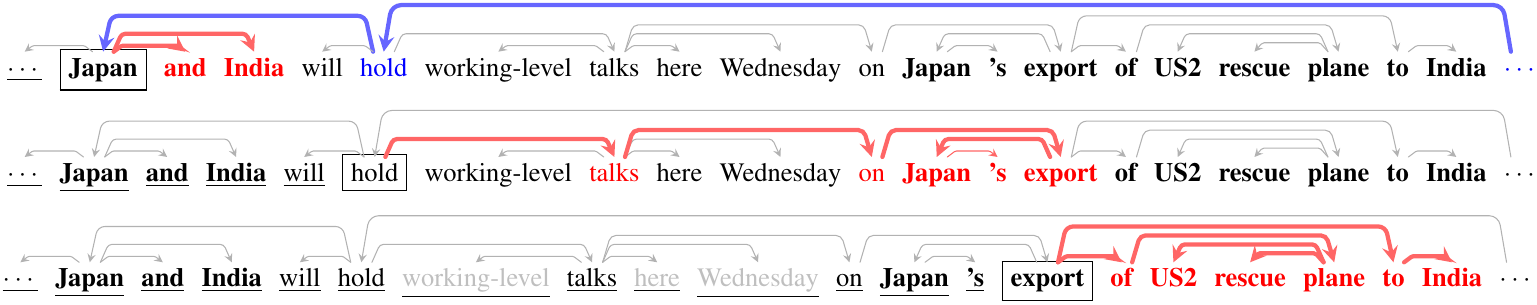}
\caption{An example sentence and its dependency tree during the decoding process.
The gray words represent deleted words, and the words in black frames are currently decoded words.
Already decoded words are underlined.
The tracking of parent nodes is represented as blue edges, and that of child nodes is represented as red edges.
The bold words represent the important words in this sentence.\label{fig:exp}}
\end{figure*}

However, since their method focuses only on parent words, important child words of the currently decoded word would be lost in compressed sentences. That is, in Seq2Seq models, because the decoder unidirectionally compresses sentences,  it cannot usually explicitly capture the relationships between decoded words and unseen words which will be decoded in the future time steps.
As the result, to avoid producing ungrammatical sentences, the decoder sometimes drops important words in compressing sentences.
To solve the problem, we need to track both parent and child words to capture unseen important words that will be decoded in the future time steps.

Fig.\ref{fig:exp} shows an example of sentence compression\footnote{This sentence actually belongs to the test set of the Google sentence compression dataset (\url{https://github.com/google-research-datasets/sentence-compression}).} that needs to track both parent and child words.
Since the input sentence mentions the export of the plane between two countries, we have to retain the name of the plane, import country and export country in the compressed sentence.

When the decoder reads ``Japan'', it should recursively track both the parent and child words of ``Japan''. Then, it can decide to retain ``hold'' that is the parent of ``Japan'' and the syntactic head of the sentence. By retaining ``hold'' in the compressed sentence, it can also retain ``Japan'', ``and'' and ``India'' because these are the child and grandchild of ``hold’' (the top case in Fig.\ref{fig:exp}).

When the decoder reads ``hold'', it should find the important phrase ``Japan's
export'' by recursively tracking child words from ``hold''. The tracking also supports the decoder for retaining ``talks'' and ``on'' to produce grammatical compression (the middle case).

When the decoder reads ``export'', it should track child words to find the important phrase ``US2 rescue plane'' and retain ``of'' for producing grammatical compression (the bottom case). 

\begin{figure}[t]
    \centering
    \includegraphics[width=0.9\columnwidth]{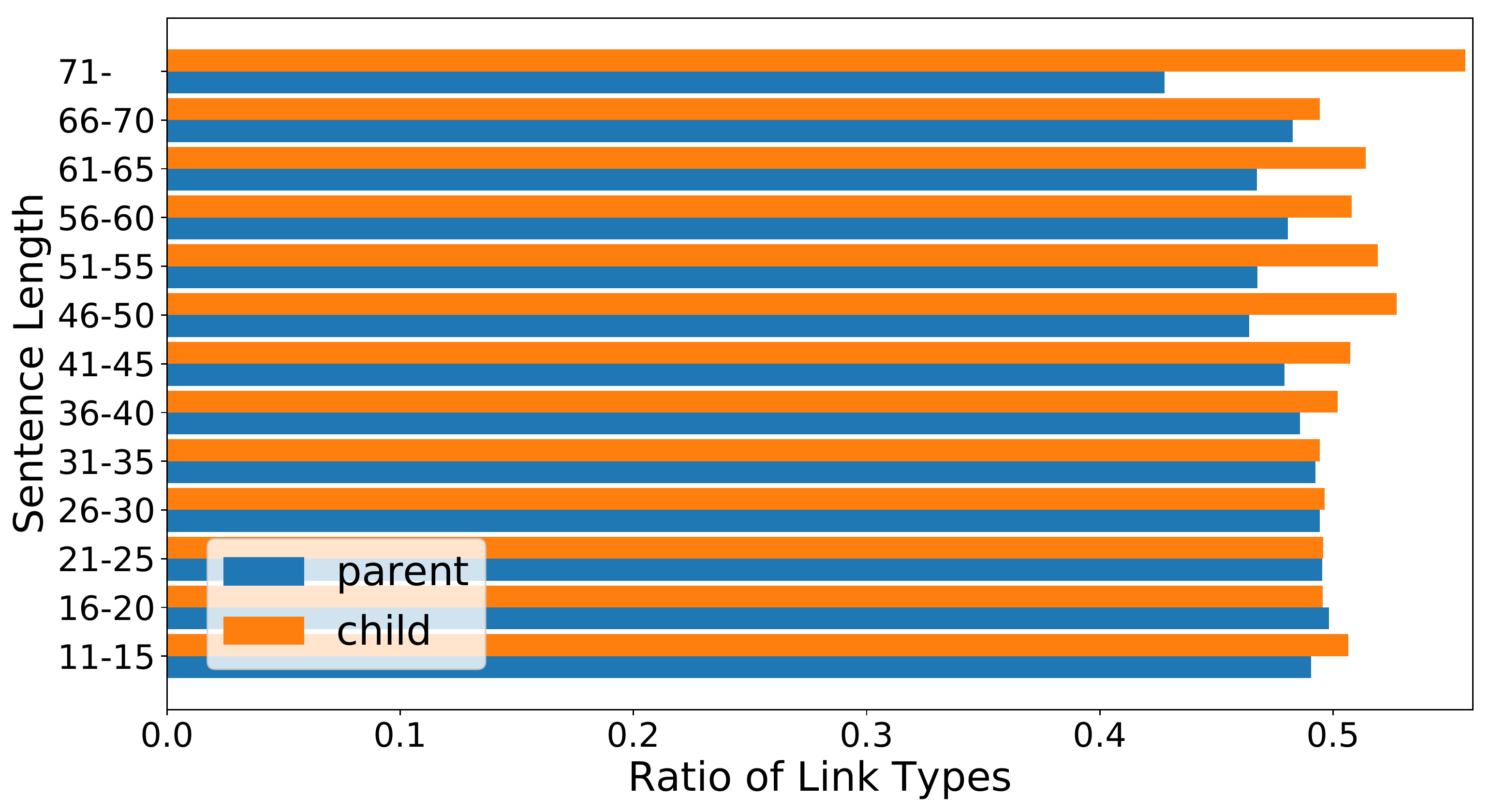}
    \caption{The proportion of words retained later that are linked from right to the retained words in the summary as a parent or a child word in the left-to-right decoding.\label{fig:length}}
\end{figure}

Note that a decoder that tracks only parent words cannot
find the important phrases or produce grammatical compression in this example.
Furthermore, Fig.\ref{fig:length} shows that tracking only parent words is not sufficient for Seq2Seq models to cover explicitly important words which will be decoded in the future time steps, especially in long sentences\footnote{This statistic is calculated on the gold compression and its dependency parse result, which are contained in the training set of the Google sentence compression dataset.}.

To incorporate this idea into Seq2Seq models, we propose \textit{syntactically look-ahead attention network} (SLAHAN), which can generate informative summaries by considering important words that will be decoded in the future time steps by explicitly tracking both parent and child words during decoding. 
The recursive tracking of dependency trees in SLAHAN is represented as attention distributions and is jointly learned with generating summaries to alleviate the effect of parse errors.
Furthermore, to avoid the bias of parent and child words, the importance of the information from recursively tracked parent and child words is automatically decided with our gate module.

The evaluation results on the Google sentence compression
dataset showed that SLAHAN achieved the best kept-token-based-F1, ROUGE-1, ROUGE-2 and ROUGE-L scores of 85.5, 79.3, 71.3 and 79.1, respectively.
SLAHAN also improved the summarization performance on longer sentences.
In addition, the human evaluation results showed that SLAHAN improved informativeness without losing readability.

\section{Our Base Seq2Seq Model\label{sec:base}}

Sentence compression is a kind of text generation task.
However, it can also be considered as a sequential tagging task, where given a sequence of input tokens $\mathbf{x} = (x_{0}, ..., x_{n})$, a sentence summarizer predicts an output label $y_t$ from specific labels (``keep'', ``delete'' or ``end of a sentence'') for each corresponding input token $x_t$~($1\leq t\leq n$). Note that $x_{0}$ is the start symbol of a sentence.

To generate a grammatically correct summary, we choose Seq2Seq models as our base model.
For constructing a robust baseline model, we introduce recently proposed contextualized word embeddings such as ELMo \cite{peters-etal-2018-deep} and BERT \cite{devlin2018bert} into the sentence compression task.
As described later in our evaluation results, this baseline exceeds the state-of-the-art $F_{1}$ scores reported by \newcite{zhao-etal-2018-language}.

Our base model consists of embedding, encoder, decoder, and output layers.
In the embedding layer, the model extracts features from an input token $x_{i}$ as a vector $e_{i}$ as follows:
\begin{equation}
\scalebox{0.8}{$
e_{i} = {\Vert}_{j=1}^{|\mathbf{F}|}F_{i,j},
\label{eq:embed}
$}
\end{equation}
where $\Vert$ represents the vector concatenation, $F_{i,j}$ is
a vector of the $j$-th feature for token $x_{i}$, and $|\mathbf{F}|$ is the number of features (at most 3). We choose features from GloVe \cite{pennington2014glove}, ELMo or BERT vectors.
Because ELMo and BERT have many layers, we treat their weighted sum as $F_{i,j}$ as follows:
\begin{equation}
\scalebox{0.8}{$
\begin{aligned}
F_{i,j} =& {\textstyle\sum}_{k=1}^{|\mathbf{L}|}\psi_{j,k} \cdot L_{i,j,k},&\\
\psi_{j,k} =& exp(\phi_{j,k} \cdot L_{i,j,k}) / {\textstyle\sum}_{l=1}^{|\mathbf{L}|}exp(\phi_{j,l} \cdot L_{i,j,l}),&
\end{aligned}
$}
\end{equation}
where $L_{i,j,k}$ represents the
$k$-th layer of the $j$-th feature for the token $x_{i}$, and $\phi_{j,k}$ is the weight vector for
the $k$-th layer of the $j$-th feature. 
In BERT, to align the input token and the output label, we treat the average of sub-word vectors as a single word vector.

The encoder layer first converts $e_{i}$ into a hidden state $\overrightarrow{h}_{i} = LSTM_{\overrightarrow{\theta}}(\overrightarrow{h}_{i-1}, e_{i})$ by using forward-LSTM, and $\overleftarrow{h}_{i}$ is calculated similarly by using backward-LSTM.
Secondly, $\overrightarrow{h}_{i}$ and $\overleftarrow{h}_{i}$ are concatenated as $h_{i}= [\overrightarrow{h}_{i}, \overleftarrow{h}_{i}]$.
Through this process, the encoder layer converts the embedding $\mathbf{e}$ into a sequence of hidden states:
\begin{equation}
\scalebox{0.8}{$
\mathbf{h} = (h_{0}, ..., h_{n}).
$}
\label{eq:enc_hidden}
\end{equation}
The final state of the backward LSTM $\overleftarrow{h}_{0}$ is inherited by the decoder as its initial state.

At time step $t$, the decoder layer encodes the concatenation of a 3-bit one-hot vector determined by the predicted label $y_{t-1}$, the final hidden state $d_{t-1}$ (which we will explain later), 
and the token embedding $e_{t}$ into the decoder hidden state $\overrightarrow{s}_{t}$, by using a forward-LSTM.

The output layer predicts an output label probability as follows:
\begin{equation}
    \scalebox{0.8}{$
    \begin{aligned}
    P(y_{t} \mid y_{<t}, \mathbf{x})=& softmax(W_{o} d_{t}) \cdot \delta_{y_{t}},\\
    d_{t} =& tanh(W_{d}[ h_{t}, \overrightarrow{s}_{t} ] + b_{d}),
    \label{eqn:lasth}
    \end{aligned}
    $}
\end{equation}
where $W_{d}$ is the weight matrix, $b_{d}$ is the bias term, $W_{o}$ is the weight matrix of the softmax layer, and $\delta_{y_{t}}$ is the
binary vector where the $y_{t}$-th element is set to $1$ and the other elements are set to $0$.

\begin{figure*}
    \centering
    \includegraphics[width=0.95\textwidth]{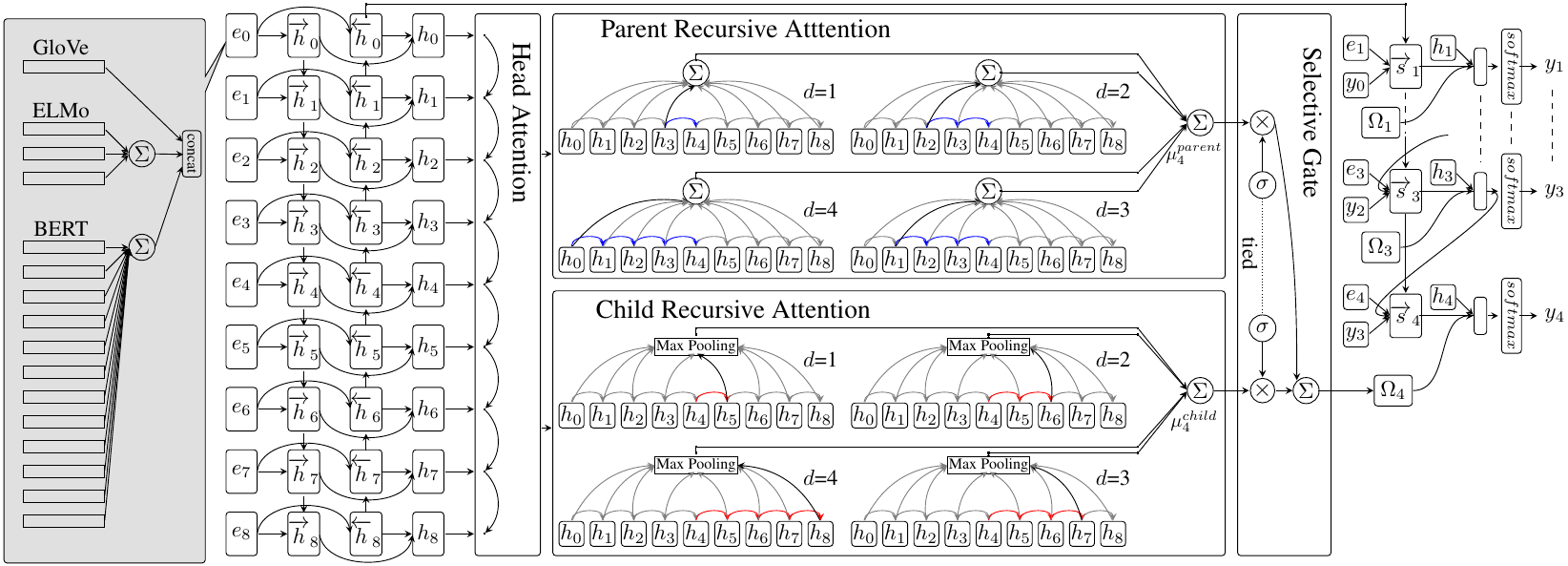}
    \caption{The entire network structure of our Syntactically Look-Ahead Attention Network (SLAHAN).}
    \label{fig:my_label}
\end{figure*}

\section{Syntactically Look-Ahead Attention Network}
\label{sec:slahan}
In this section, we first explain the graph representation for a dependency tree that is used in SLAHAN and then introduce its entire network structure and the modules inside it.
Both the graph representation and network parameters are jointly updated, as described in the later section.

\subsection{Graph Representation of Dependency Relationships}
We explain the details of our representation for tracking parent and child words from a word in a dependency tree.
As described in \newcite{hashimoto-tsuruoka:2017:EMNLP2017}, a dependency relationship can be represented as a weighted graph.
Given sentence $\mathbf{x} = (x_{0}, ..., x_{n})$, the parent of each word $x_{j}$ is selected from $\mathbf{x}$. We treat $x_{0}$ as a root node.
We represent the probability of $x_{j}$ being the parent of $x_{t}$ in $\mathbf{x}$ as $P_{head}(x_{j}|x_{t},\mathbf{x})$.
By using $P_{head}(x_{j}|x_{t},\mathbf{x})$, \newcite{kamigaito-etal-2018-higher} show that $\alpha_{d,t,j}$, a probability of $x_{j}$ being the $d$-th order parent of $x_{t}$, is calculated as follows:
\begin{equation}
\scalebox{0.8}{$
\alpha_{d,t,j} = \left\{\begin{array}{ll}
{\textstyle\sum}_{k=1}^{n}\alpha_{d-1,t,k} \cdot \alpha_{1,k,j} & (d \! > \! 1) \\
P_{head}(x_{j}|x_{t},\mathbf{x}) & (d \! = \! 1)
\end{array} \right..
$}
\label{eqn:recatt}
\end{equation}
Because the 1st line of Eq.(\ref{eqn:recatt}) is a definition of matrix multiplication,
by using a matrix $A^{d}$, which satisfies $A^{d}_{j,t} = \alpha_{d,t,j}$,
Eq.(\ref{eqn:recatt}) is reformulated as follows:
\begin{equation}
    \scalebox{0.8}{$
    A^{d} = A^{d-1}A^{1}.
    $}
    \label{eq:recatt:matmul}
\end{equation}
We call $A^{d}$ the \textit{$d$-th parent graph} hereafter.

We expand Eq.(\ref{eq:recatt:matmul}) to capture $d$-th child words of a word $x_{j}$.
At first, we define $P_{child}(x_{t}|x_{j},\mathbf{x})$, the probability of $x_{t}$ being a child of $x_{j}$ in $\mathbf{x}$, $P_{\mathbf{x}}(x_{j}=p)$, the probability of $x_{j}$ being a parent word in $\mathbf{x}$, $P_{\mathbf{x}}(x_{t}=c)$, the probability of $x_{t}$ being a child word in $\mathbf{x}$, and $P_{\mathbf{x}}(x_{j}, x_{t})$, the  probability of $x_{j}$ and $x_{t}$ having a link in $\mathbf{x}$.
Assuming the probability of words having a link is independent of each other, the following equations are satisfied: 
\begin{equation}
\scalebox{0.8}{$
\begin{aligned}
    P_{\mathbf{x}}(x_{j}, x_{t}) &= P_{child}(x_{t}|x_{j},\mathbf{x}) \cdot P_{\mathbf{x}}(x_{j}=p),\\
    P_{\mathbf{x}}(x_{j}, x_{t}) &= P_{head}(x_{j}|x_{t},\mathbf{x}) \cdot P_{\mathbf{x}}(x_{t}=c).
\end{aligned}
$}    
\end{equation}
This can be reformulated as follows:
\begin{equation}
\scalebox{0.8}{$
    \!\!\!\!\!\!\!\!P_{child}(x_{t}|x_{j},\mathbf{x}) \!\!=\!\! P_{head}(x_{j}|x_{t},\mathbf{x}) \! \cdot \! P_{\mathbf{x}}(x_{t}\!=\!c) / P_{\mathbf{x}}(x_{j} = p).\!\!\!\!\!\!
$}    
\end{equation}
Here, $P_{\mathbf{x}}(x_{t}=c)$ is always $1$ because of the dependency tree definition, and in this  formulation, we treat $x_{j}$ as a parent;
thus, $P_{\mathbf{x}}(x_{j}=p)$ is a constant value.
Therefore, we can obtain the following relationship:
\begin{equation}
    \scalebox{0.8}{$
    P_{child}(x_{t}|x_{j},\mathbf{x}) \propto P_{head}(x_{j}|x_{t},\mathbf{x}).
    $}
    \label{eq:prop}
\end{equation}
Based on Eq.(\ref{eq:prop}), we can define $\beta_{d,t,j}$, the strength of $x_{j}$ being the $d$-th order child of $x_{t}$, as follows:
\begin{equation}
\scalebox{0.8}{$
\beta_{d,t,j} = \left\{\begin{array}{ll}
{\textstyle\sum}_{k=1}^{n}\beta_{d-1,t,k} \cdot \beta_{1,k,j} & (d \! > \! 1) \\
P_{head}(x_{t}|x_{j},\mathbf{x}) & (d \! = \! 1)
\end{array} \right..
$}
\label{eqn:recatt:trans}
\end{equation}
Similar to Eq.(\ref{eqn:recatt}), by using a matrix $B^{d}$, which satisfies $B^{d}_{j,t} = \beta_{d,t,j}$,
Eq.(\ref{eqn:recatt:trans}) is reformulated as follows:
\begin{equation}
    \scalebox{0.8}{$
    B^{d} = B^{d-1}B^{1}.
    $}
    \label{eqn:recatt:trans:matmul}
\end{equation}
We call $B^{d}$ the \textit{$d$-th child graph} hereafter.
Note that from the definition of the 2nd lines in Eq.(\ref{eqn:recatt}) and Eq.(\ref{eqn:recatt:trans}), $A^{1}$ and $B^{1}$ always satisfy $B^{1}_{tj} = A^{1}_{jt}$. This can be reformulated as $B^{1}=(A^{1})^{T}$.
Furthermore, from the definition of the transpose of a matrix, we can obtain the following formulation:
\begin{equation}
    \scalebox{0.8}{$
    \!\!\!\!\!\!\!\! B^{d} = B^{1}B^{1} \cdots B^{1} = (A^{1})^{T}(A^{1})^{T} \cdots (A^{1})^{T} = (A^{d})^{T}. \!\!\!\!
    $}
\end{equation}
Thus, once we calculate Eq.(\ref{eq:recatt:matmul}), we do not need to compute Eq.(\ref{eqn:recatt:trans:matmul}) explicitly.
Therefore, letting $d$ be a dimension size of hidden vectors, the computational cost of SLAHAN is $O(n^2d^2)$, similar to \newcite{kamigaito-etal-2018-higher}.
This is based on the assumption that $d$ is larger than $n$ in many cases.
Note that the computational cost of the base model is $O(nd^2)$. 

\subsection{Network Structure}
Fig.\ref{fig:my_label} shows the entire structure of SLAHAN.
It is constructed on our base model, as described in the previous section.
After encoding the input sentence, the hidden states are passed to our network modules.
The functions of each module are as follows:\\
{\bf Head Attention} module makes a dependency graph of a sentence by calculating the probability of $x_{j}$ being the parent of $x_{t}$ based on $h_{j}$ and $h_{t}$ in Eq.(\ref{eq:enc_hidden}) for each $x_{t}$. \\
{\bf Parent Recursive Attention} module calculates $d$-th parent graph $A^{d}$ and extracts a weighted sum of important hidden states $\mu^{parent}_{t}$ from $\mathbf{h}$ in Eq.(\ref{eq:enc_hidden}) based on $\alpha_{d,t,j}$ ($=A^{d}_{j,t}$) for each decoder time step $t$. \\
{\bf Child Recursive Attention} module uses $d$-th child graph $B^{d}$ to extract $\mu^{child}_{t}$, a weighted sum of important hidden states from $\mathbf{h}$ in Eq.(\ref{eq:enc_hidden}) based on $\beta_{d,t,j}$ ($=B^{d}_{j,t}$) for each decoder time step $t$. \\
{\bf Selective Gate} module supports the decoder to capture important words that will be decoded in the future by calculating $\Omega_{t}$, the weighted sum of $\mu^{parent}_{t}$ and $\mu^{child}_{t}$, based on the current context. $\Omega_{t}$ is inherited to the decoder for deciding the output label $y_{t}$. 

The details of each module are described in the following subsections.

\subsubsection{Head Attention}
Similar to \newcite{zhang-cheng-lapata:2017:EACLlong}, we calculate $P_{head}(x_{j}|x_{t},\mathbf{x})$ as follows:
\begin{equation}
    \scalebox{0.8}{$
\begin{aligned}
\!\!\!\!\!\!&\!\!\!\!\!&\!\!\!\!\!P_{head}(x_{j}|x_{t},\mathbf{x}) \!&=\! softmax(g(h_{j'}, h_{t})) \cdot \delta_{x_{j}},\\
\!\!\!\!\!\!&\!\!\!\!\!&\!\!\!\!\!g(h_{j'}, h_{t}) \! &= \! v_{a}^{T} \cdot tanh(U_{a}\cdot h_{j'} + W_{a} \cdot h_{t}),
\end{aligned}
$}
\end{equation}
where $v_{a}$, $U_{a}$ and $W_{a}$ are weight matrices of $g$.
In a dependency tree, the {\it root} has no parent, and a token does not depend on itself.
In order to satisfy these rules, we impose the following constraints on $P_{head}(x_{j}|x_{t},\mathbf{x})$:
\begin{equation}
\scalebox{0.8}{$
P_{head}(x_{j}|x_{t},\mathbf{x}) = \left\{\begin{array}{ll}
1 & (t = 0 \land j = 0) \\
0 & (t = 0 \land j > 0) \\
0 & (t \not= 0 \land t = j).
\end{array} \right.
$}
\label{eqn:recatt:const}
\end{equation}
The 1st and 2nd lines of Eq.(\ref{eqn:recatt:const}) represent the case where the parent of {\it root} is also {\it root}.
These imply that {\it root} does not have a parent.
The 3rd line of Eq.(\ref{eqn:recatt:const}) prevents a token from depending on itself.
In the training phase, $P_{head}(x_{j}|x_{t},\mathbf{x})$ is jointly learned with output label probability $P(\mathbf{y} \mid \mathbf{x})$, as described in the objective function section.

\subsubsection{Parent Recursive Attention}

The parent recursive attention module recursively calculates $\alpha_{d,t,j}$ by using $P_{head}(x_{j}|x_{t},\mathbf{x})$ based on Eq.(\ref{eqn:recatt}).
The calculated $\alpha_{d,t,j}$ is used to weight the bi-LSTM hidden layer $\mathbf{h}$ as follows:
\begin{equation}
\scalebox{0.8}{$
\gamma_{d,t} = {\textstyle\sum}_{k=j}^{n} \alpha_{d,t,j} \cdot h_{j}.
$}
\label{eqn:rec:par:softmax}
\end{equation}

To select suitable dependency order $d$ for the input sentence, $\gamma_{d,t}$ is  further weighted and summed to $\mu^{parent}_{t}$ by using weighting parameter $\eta_{d,t}$, according to the current context as follows:
\begin{equation}
\scalebox{0.8}{$
\begin{aligned}
c_{t} &= [ \overleftarrow{h}_{0}, \overrightarrow{h}_{n}, h_{t}, \overrightarrow{s}_{t} ],\\
    \eta_{d,t} &= softmax(\gamma_{d,t}W^{parent}_{d}c_{t}) \cdot \delta_{d}, \\
    \mu^{parent}_{t} &= {\textstyle\sum}_{d \in \mathbf{d}}\eta_{d,t} \cdot \gamma_{d,t},
\label{eqn:rec:parent:order:softmax}
\end{aligned}
$}
\end{equation}
where $W^{parent}_{d}$ is the weight matrix, $\mathbf{d}$ is the group of dependency orders, and $c_{t}$ is the vector representing the current context.

\subsubsection{Child Recursive Attention}

The child recursive attention module weights the bi-LSTM hidden layer $\mathbf{h}$ based on $d$-th child graph $B^{d}$.
Unlike the parent recursive attention module, $B^{d}$ is not a probability, and a word sometimes has more than two children.
For that reason, we use max-pooling rather than the attention distribution in Eq.(\ref{eqn:rec:par:softmax}).
In the child recursive attention module, the bi-LSTM hidden layer $\mathbf{h}$ is weighted by $\beta_{d,t,j}$ and then pooled as follows:
\begin{equation}
\scalebox{0.8}{$
\rho_{d,t} = MaxPool(\parallel_{k=j}^{n} (\beta_{d,t,j} \cdot h_{j})^{T}).
$}
\end{equation}

To select suitable dependency order $d$ for the input sentence, $\rho_{d,t}$ is further weighted and summed to $\mu^{child}_{t}$ by using weighting parameter $\eta_{d,t}$, according to the current context as follows:
\begin{equation}
\scalebox{0.8}{$
\begin{aligned}
    \eta_{d,t} &= softmax(\rho_{d,t}W^{child}_{d}c_{t}) \cdot \delta_{d}, \\
    \mu^{child}_{t} &= {\textstyle\sum}_{d \in \mathbf{d}}\eta_{d,t} \cdot \rho_{d,t},
\label{eqn:rec:child:order:softmax}
\end{aligned}
$}
\end{equation}
where $W^{child}_{d}$ is the weight matrix.

\subsubsection{Selective Gate}
This module calculates $\Omega_{t}$, a weighted sum of parent information $\mu^{parent}_{t}$ and child information $\mu^{child}_{t}$. The weight is decided by a gate $z_{t}$ by considering whether $\mu^{parent}_{t}$ or $\mu^{child}_{t}$ is more important in the current context.
Specifically, $\Omega_{t}$ is calculated as follows:
\begin{equation}
\scalebox{0.8}{$
\begin{aligned}
    \Omega_{t} &= z_{t} \circ \mu^{parent}_{t} + (1-z_{t}) \circ \mu^{child}_{t}, \label{eqn:gate}\\
    z_{t} &= \sigma(W_{z}[\mu^{parent}_{t},\mu^{child}_{t},c_{t}]),
\end{aligned}
$}
\end{equation}
where $\circ$ is the element-wise product, $\sigma$ is the sigmoid function, and $W_{z}$ is the weight matrix.
Then, $d_{t}$ in Eq.(\ref{eqn:lasth}) is replaced by a concatenated vector $d'_{t} = [ h_{t}, \Omega_{t}, \overrightarrow{s}_{t} ]$; furthermore, instead of $d_{t}$, $d'_{t}$ is also fed to the decoder input at $t+1$. 

\subsection{Objective Function}
\label{sec:objfunc}
To alleviate the effect of parse errors, we jointly update dependency parent probability $P_{head}(x_{j}|x_{t})$ and label probability $P(\mathbf{y}|\mathbf{x})$ \cite{kamigaito-etal-2017-supervised}.
We denote the existence of an edge between parent word $w_{j}$ and child word $w_{t}$ on a dependency tree as $a_{t,j}=1$.
In contrast, we denote the absence of an edge as $a_{t,j}=0$. By using these notations, our objective function is defined as follows:
\begin{equation}
\scalebox{0.8}{$
-log P(\mathbf{y}|\mathbf{x}) -\lambda \cdot {\textstyle\sum}_{j=1}^{n} {\textstyle\sum}_{t=1}^{n} a_{t,j} \cdot log \alpha_{1,t,j},
$}
\label{eqn:sup}
\end{equation}
where $\lambda$ is a hyper-parameter balancing the importance of output labels and parse trees in training steps.
To investigate the importance of the syntactic information, we used $\lambda=1.0$ for the \textit{with syntax (w/ syn)} setting and $\lambda=0$ for the \textit{without syntax (w/o syn)} setting.

\section{Experiments}
For comparing our proposed models with the baselines, we conducted both automatic and human evaluations. The following subsections describe the evaluation details.
\subsection{Settings}
\subsubsection{Datasets}

We used the Google sentence compression dataset (Google dataset) \cite{filippova-altun:2013:EMNLP} for our evaluations. To evaluate the performances on an out-of-domain dataset, we also used the Broadcast News Compression Corpus (BNC Corpus)\footnote{\url{https://www.jamesclarke.net/research/resources}}.
The setting for these datasets is as follows:\\
{\bf Google dataset:} Similar to the previous researches \cite{filippova-EtAl:2015:EMNLP,Tran:2016:EAN:3011077.3011111,wang-EtAl:2017:Long4,kamigaito-etal-2018-higher,zhao-etal-2018-language}, we used the first 1,000 sentences of {\sl comp-data.eval.json} as the test set.
We used the last 1,000 sentences of {\sl comp-data.eval.json} as our development set.
Following recent researches \cite{kamigaito-etal-2018-higher,zhao-etal-2018-language}, we used all 200,000 sentences in {\sl sent-comp.train*.json} as our training set.
We also used the dependency trees contained in this dataset.

To investigate the summarization performances on long sentences, we additionally performed evaluations on 417 sentences that are longer than the average sentence length ($=27.04$) in the test set.\\
{\bf BNC Corpus:} This dataset contains spoken sentences and their summaries created by three annotators. 
To evaluate the compression performances on long sentences in the out-of-domain setting, we treated sentences longer than the average sentence length, $19.83$, as the test set (595 sentences), and training was conducted with the Google dataset.
Because this dataset does not contain any dependency parsing results, we parsed all sentences in this dataset by using the Stanford dependency parser\footnote{\url{https://nlp.stanford.edu/software/}}.
In all evaluations, we report the average scores for three annotators.

\subsubsection{Compared Models}
The baseline models are as follows. We used ELMo, BERT and GloVe vectors for all models in our experiments.\\
{\bf Tagger:} This is a bi-LSTM tagger which is used in various sentence summarization researches \cite{klerke-goldberg-sogaard:2016:N16-1,wang-EtAl:2017:Long4}.\\
{\bf LSTM:} This is an LSTM-based sentence summarizer, which was proposed by \newcite{filippova-EtAl:2015:EMNLP}.\\
{\bf LSTM-Dep:} This is an LSTM-based sentence summarizer with dependency features, called LSTM-Par-Pres in \newcite{filippova-EtAl:2015:EMNLP}.\\
{\bf Base:} Our base model explained in the 2nd section.\\
{\bf Attn:} This is an improved attention-based Seq2Seq model with ConCat attention, described in \newcite{luong-pham-manning:2015:EMNLP}. To capture the context of long sentences, we also feed input embedding into the decoder, similar to the study of \newcite{filippova-EtAl:2015:EMNLP}.\\
{\bf Parent:} This is a variant of SLAHAN that does not have the child recursive attention module. This model captures only parent words, similar to HiSAN in the study of \newcite{kamigaito-etal-2018-higher}. For the fair comparisons, we left the gate layer in Eq.(\ref{eqn:gate}).

Our proposed models are as follows:\\
{\bf SLAHAN:} This is our proposed model which is described in the 3rd section.\\
{\bf Child:} This is a variant of SLAHAN that does not have the parent recursive attention module. Similar to {\bf Parent}, we left the gate layer in Eq.(\ref{eqn:gate}).

\begin{table}[t]
    \centering
\resizebox {0.8\columnwidth} {!} {
\small
    \begin{tabular}{|l|c|c|c|c|c|c|c|}
         \hline
         Glove              & $\checkmark$ & $\checkmark$ &              & $\checkmark$ & $\checkmark$ &              &              \\
         ELMo               & $\checkmark$ & $\checkmark$ & $\checkmark$ &              &              & $\checkmark$ &              \\
         BERT               & $\checkmark$ &              & $\checkmark$ & $\checkmark$ &              &              & $\checkmark$ \\
         \hline
         $\mathbf{F_{1}}$   & $\mathbf{86.2}$ & 86.0 & 85.9 & 85.4 & 85.5 & 85.9 & 84.8 \\
         \hline
    \end{tabular}
}
    \caption{$F_{1}$ scores for \textbf{Base} with various features in the development data. The bold score represents the highest score. \label{tb:eval:dev}}
\end{table}
\begin{table*}[t]
\centering
\resizebox {0.8\textwidth} {!} {
\small
\begin{tabular}{llcccccccccccc}
\toprule
& & \multicolumn{5}{c}{\textbf{ALL}} & \multicolumn{5}{c}{\textbf{LONG}}
\\
\cmidrule(lr){3-7} \cmidrule(lr){8-12}
 & & $\mathbf{F_1}$ & \textbf{R-1} & \textbf{R-2} & \textbf{R-L} & $\mathbf{\Delta C}$ & $\mathbf{F_1}$ & \textbf{R-1} & \textbf{R-2} & \textbf{R-L} & $\mathbf{\Delta C}$ \\
\midrule
\multicolumn{2}{l}{Evaluator-LM \cite{zhao-etal-2018-language}}    & 85.0 & - & - & - & -2.7 & - & - & - & - & -\\
\multicolumn{2}{l}{Evaluator-SLM \cite{zhao-etal-2018-language}}    & 85.1 & - & - & - & -4.7 & - & - & - & - & -\\
\midrule
\multicolumn{2}{l}{Tagger}  & 85.0 & 78.1 & 69.9 & 77.9 & -3.1 & 83.0 & 75.4 & 66.8 & 74.9 & -3.1\\
\multicolumn{2}{l}{LSTM}    & 84.8 & 77.7 & 69.6 & 77.4 & -3.4 & 82.7 & 74.8 & 66.3 & 74.4 & -3.5\\
\multicolumn{2}{l}{LSTM-Dep}& 84.7 & 77.8 & 69.7 & 77.5 & -3.3 & 82.6 & 74.9 & 66.5 & 74.4 & -3.3\\
\multicolumn{2}{l}{Attn}    & 84.5 & 77.3 & 69.3 & 77.1 & -3.8 & 82.3 & 74.7 & 66.4 & 74.3 & -3.6\\
\multicolumn{2}{l}{Base}    & 85.4 & 78.5 & 70.4 & 78.2 & -2.9 & 83.4 & 75.8 & 67.4 & 75.3 & -3.0\\
Parent & w/ syn             & 85.0 & 78.3 & 70.3 & 78.1 & -2.5 & 82.8 & 75.3 & 67.0 & 74.9 & -2.9\\
Parent & w/o syn            & 85.3 & 78.3 & 70.4 & 78.1 & -3.4 & 83.3 & 75.6 & 67.3 & 75.2 & -3.4\\
\midrule
Child & w/ syn              & 85.4 & 78.8 & 70.7 & 78.5 & -2.9 & 83.0 & 75.8 & 67.3 & 75.4 & -3.0\\
Child & w/o syn             & 85.2 & 78.6 & 70.8 & 78.4 & -3.1 & 83.2 & 76.3 & 68.2 & 75.8 & -2.8\\
SLAHAN & w/ syn             & $\mathbf{85.5}$ & $\mathbf{79.3}^{\dagger}$ & $\mathbf{71.4}^{\dagger}$ & $\mathbf{79.1}^{\dagger}$ & $\mathbf{-1.5}^{\dagger}$ & 83.3 & $\mathbf{76.6}$ & 68.3 & $\mathbf{76.1}$ & $\mathbf{-1.9}^{\dagger}$\\
SLAHAN & w/o syn            & 85.4 & $78.9^{\dagger}$ & $71.0^{\dagger}$ & $78.6^{\dagger}$ & -3.0 & $\mathbf{83.6}$ & $76.5^{\dagger}$ & $\mathbf{68.5}^{\dagger}$ & $\mathbf{76.1}^{\dagger}$ & -2.9\\
\bottomrule
\end{tabular}
}
    \caption{Results on the Google dataset. $\mathbf{ALL}$ and $\mathbf{LONG}$ represent, respectively, the results for all sentences and only for long sentences (longer than average length 27.04) in the test dataset. The bold values indicate the best scores. $\dagger$ indicates that the difference of the score from the best baseline (mostly Base) is statistically significant.\footnotemark[8]} \label{tb:eval:google}
\end{table*}
\begin{table}[t]
\centering
\small
\resizebox {0.825\columnwidth} {!} {
\begin{tabular}{llccccc}
\toprule
 & & $\mathbf{F_1}$ & \textbf{R-1} & \textbf{R-2} & \textbf{R-L} & $\mathbf{\Delta C}$ \\
\midrule
\multicolumn{2}{l}{Tagger}  & 54.6 & 36.8 & 27.7 & 36.4 & -39.1\\
\multicolumn{2}{l}{LSTM}    & 54.8 & 36.6 & 28.0 & 36.2 & -39.2\\
\multicolumn{2}{l}{LSTM-Dep}& 55.1 & 36.9 & 28.2 & 36.5 & -38.8\\
\multicolumn{2}{l}{Attn}    & 54.1 & 36.1 & 27.4 & 35.6 & -39.6\\
\multicolumn{2}{l}{Base}    & 55.4 & 37.4 & 28.5 & 36.9 & -38.6\\
Parent & w/ syn             & 54.2 & 36.3 & 27.7 & 35.9 & -39.1\\
Parent & w/o syn            & 54.0 & 35.8 & 27.2 & 35.4 & -40.1\\
\midrule
Child & w/ syn              & 55.6 & 37.8 & 28.5 & 37.3 & -38.2\\
Child & w/o syn             & 54.8 & 36.7 & 28.1 & 36.3 & -39.2\\
SLAHAN & w/ syn             & $\mathbf{57.7}^{\dagger}$ & $\mathbf{40.1}^{\dagger}$ & $\mathbf{30.6}^{\dagger}$ & $\mathbf{39.6}^{\dagger}$ & $\mathbf{-35.9}^{\dagger}$\\
SLAHAN & w/o syn            & 54.6 & 36.4 & 27.8 & 36.0 & -39.5\\
\bottomrule
\end{tabular}}
    \caption{Results on the BNC Corpus. $\dagger$ indicates the same as in Table \ref{tb:eval:google}.\label{tb:eval:bnc}}
\end{table}

\subsubsection{Model Parameters}

We used GloVe ({\sl glove.840B.300d}), 3-layers of ELMo and 12-layers of BERT ({\sl cased\_L-12\_H-768\_A-12}) as our features.
We first investigated the best combination of GloVe, ELMo, and BERT vectors as shown in Table \ref{tb:eval:dev}.
Following this result, we used the combination of all of GloVe, ELMo and BERT for all models.

The dimensions of the LSTM layer and the attention layer were set to 200.
The depth of the LSTM layer was set to 2.
These sizes were based on the setting of the LSTM NER tagger with ELMo in the study of \newcite{peters-etal-2018-deep}.
All parameters were initialized with \newcite{glorot2010understanding}'s method.
For all methods, we applied Dropout \cite{srivastava2014dropout} to the input of the LSTM layers.
All dropout rates were set to 0.3.
We used Adam \cite{DBLP:journals/corr/KingmaB14} with an initial learning rate of 0.001 as our optimizer.
All gradients were averaged by the number of sentences in each mini-batch.
The clipping threshold value for the gradients was set to 5.0.
The maximum training epoch was set to 20.
We used $\{1,2,3,4\}$ as $\mathbf{d}$ in Eq.(\ref{eqn:rec:parent:order:softmax}) and  Eq.(\ref{eqn:rec:child:order:softmax}). 
The maximum mini-batch size was set to 16, and the order of mini-batches was shuffled at the end of each training epoch.
We adopted early stopping to the models based on maximizing per-sentence accuracy (i.e., how many summaries are fully reproduced) of the development data set.

To obtain a compressed sentence, we used greedy decoding, following the previous research \cite{kamigaito-etal-2018-higher}.
We used Dynet \cite{neubig2017dynet} to implement our neural networks\footnote{Our implementation is publicly available on GitHub at \newline\url{https://github.com/kamigaito/slahan}.}.

\subsection{Automatic Evaluation}

\subsubsection{Evaluation Metrics}

In the evaluation, we used kept-token-based-F$_{1}$ measures ($\mathbf{F_{1}}$) for comparing to the previously reported scores.
In this metric, precision is defined as the ratio of kept tokens that overlap with the gold summary, and recall is defined as the ratio of tokens in the gold summary that overlap with the system output summary.
For more concrete evaluations, we additionally used ROUGE-1 (\textbf{R-1}), ROUGE-2 (\textbf{R-2}), and ROUGE-L (\textbf{R-L}) \cite{lin-och:2004:ACL}\footnote{We used the ROUGE-1.5.5 script with option ``-n 2 -m -d -a''.} with limitation by reference byte lengths\footnote{If a system output exceeds the reference summary byte length, we truncated the exceeding tokens.} as evaluation metrics.
We used $\Delta C = system\ compression\ ratio - gold\ compression\ ratio$ \cite{kamigaito-etal-2018-higher} to evaluate how close the compression ratio of system outputs was to that of gold compressed sentences.
Note that the gold compression ratios of all the sentences and the long sentences in the Google test set are respectively $43.7$ and $32.4$. Those of all the sentences and the long sentences in the BNC corpus are respectively $76.3$ and $70.8$.
We used the macro-average for all reported scores.
All scores are reported as the average scores of three randomly initialized trials.

\subsubsection{Results}

Table \ref{tb:eval:google} shows the evaluation results on the Google dataset. 
\textbf{SLAHAN} achieved the best scores on both all the sentences and the long sentences. Through these gains, we can understand that \textbf{SLAHAN} successfully captures important words by tracking both parent and child words.
\textbf{Child} achieved better scores than \textbf{Parent}. This result coincides with our investigation that tracking child words is important especially for long sentences, as shown in Fig.\ref{fig:length}. 
We can also observe the score of \textbf{SLAHAN w/o syn} is comparable to that of \textbf{SLAHAN w/ syn}.
This result indicates that dependency graphs can work on the in-domain dataset without relying on given dependency parse trees.
\begin{table}[!t]
\centering
\small
\resizebox {0.6\columnwidth} {!} {
\begin{tabular}{lcccc}
\toprule
& \multicolumn{2}{c}{$\mathbf{Read}$} & \multicolumn{2}{c}{$\mathbf{Info}$} \\
\midrule
Tagger       & 3.90 & \hspace{-3mm}(73.4) & \hspace{-1mm}3.79 & \hspace{-3mm}(72.9) \\
Base        & 3.86 & \hspace{-3mm}(72.4) & \hspace{-1mm}3.80 & \hspace{-3mm}(73.6) \\
Parent w/ syn & 3.82 & \hspace{-3mm}(70.5) & \hspace{-1mm}3.77 & \hspace{-3mm}(71.5) \\
\midrule
Child w/ syn & $\mathbf{3.94}$ & \hspace{-3mm}($\mathbf{75.8}$) & \hspace{-1mm}$3.85^{\dagger}$ & \hspace{-3mm}(74.9) \\
SLAHAN w/ syn & 3.91 & \hspace{-3mm}(74.8) & \hspace{-1mm}$\mathbf{3.90}^{\dagger}$ & \hspace{-3mm}($\mathbf{77.9}^{\dagger}$) \\
\bottomrule
\end{tabular}
}
    \caption{Results of the human evaluation.
    The numbers in parentheses are the percentages of over four ratings. $\dagger$ indicates the same as in Table \ref{tb:eval:google}. \label{tb:eval:human}}
\end{table}

We also show the evaluation results on the BNC corpus, the out-of-domain dataset, in Table \ref{tb:eval:bnc}.
We can clearly observe that \textbf{SLAHAN w/ syn} outperforms other models for all metrics.
Comparing between \textbf{Base}, \textbf{Parent}, \textbf{Child} and \textbf{SLAHAN}, we can understand that \textbf{SLAHAN w/ syn} captured important words during the decoding step even in the BNC corpus.
The remarkable performance of \textbf{SLAHAN w/ syn} supports the effectiveness of explicit syntactic information.
That is, in the out-of-domain dataset, the dependency graph learned with
implicit syntactic information obtained lower scores than
that learned with explicit syntactic information.
The result agrees with the findings of the previous research \cite{wang-EtAl:2017:Long4}.
From these results, we can conclude that SLAHAN is effective even for both long and out-of-domain sentences.
\begin{table}[t]
    \small
    \centering
    \resizebox {0.9\columnwidth} {!} {
    \begin{tabular}{p{1.1\columnwidth}}
        \toprule
         {\bf Input}: British mobile phone giant Vodafone said Tuesday it was seeking regulatory approval to take full control of its Indian unit for \$ 1.65 billion , after New Delhi relaxed foreign ownership rules in the sector .\\
         {\bf Gold}: Vodafone said it was seeking regulatory approval to take full control of its Indian unit .\\
         {\bf Base}: Vodafone said it was seeking regulatory approval to take control of its unit .\\
         {\bf Parent w/ syn}: Vodafone said it was seeking approval to take full control of its Indian unit .\\
         {\bf Child w/ syn}: Vodafone said it was seeking regulatory approval to take control of its Indian unit . \\
         {\bf SLAHAN w/ syn}: Vodafone said it was seeking regulatory approval to take full control of its Indian unit .\\
        \midrule
         {\bf Input}: Broadway 's original Dreamgirl Jennifer Holliday is coming to the Atlanta Botanical Garden for a concert benefiting Actor 's Express .\\
         {\bf Gold}: Broadway 's Jennifer Holliday is coming to the Atlanta Botanical Garden .\\
         {\bf Base}: Jennifer Holliday is coming to the Atlanta Botanical Garden .\\
         {\bf Parent w/ syn}: Broadway 's Jennifer Holliday is coming to the Atlanta Botanical Garden .\\
         {\bf Child w/ syn}: Jennifer Holliday is coming to the Atlanta Botanical Garden .\\
         {\bf SLAHAN w/ syn}: Broadway 's Jennifer Holliday is coming to the Atlanta Botanical Garden . \\
        \midrule
         {\bf Input}: Tokyo , April 7 Japan and India will hold working-level talks here Wednesday on Japan 's export of US2 rescue plane to India , Japan 's defence ministry said Monday .\\
         {\bf Gold}: Japan and India will hold talks on Japan 's export of US2 rescue plane to India .\\
         {\bf Base}: Japan and India will hold talks Wednesday on export of plane to India .\\
         {\bf Parent w/ syn}: Japan and India will hold talks on Japan 's export plane . \\
         {\bf Child w/ syn}: Japan and India will hold talks on Japan 's export of US2 rescue plane to India . \\
         {\bf SLAHAN w/ syn}: Japan and India will hold talks on Japan 's export of plane to India .\\
        \bottomrule
    \end{tabular}
    }
    \caption{Example compressed sentences.}
    \label{tb:analysis}
\end{table}
\footnotetext[8]{We used paired-bootstrap-resampling \cite{koehn-2004-statistical} with 1,000,000 random samples ($p<0.05$).}

\subsection{Human evaluation}
In the human evaluation, we compared the models\footnote{We chose the models that achieved the highest $F_{1}$ scores in the development set from the three trials.} that achieved the top five \textbf{R-L} scores in the automatic evaluation.
We filtered out sentences whose compressions are the same for all the models and selected the first 100 sentences from the test set of the Google dataset.
Those sentences were evaluated for both readability (\textbf{Read}) and informativeness (\textbf{Info}) by twelve raters, who were asked to rate them in a five-point Likert scale, ranging from one to five for each metric.
To reduce the effect by outlier rating, we excluded raters with the highest and lowest average ratings. Thus, we report the average rates of the ten raters.

Table \ref{tb:eval:human} shows the results. {\bf SLAHAN w/ syn} and {\bf Child w/ syn} improved informativeness without losing readability,  compared to the baselines. These improvements agreed with the automatic evaluation results.

\section{Analysis}
In Table \ref{tb:eval:dev}, BERT underperforms ELMo and GloVe.
Recently, \newcite{lin-etal-2019-unified} reported that ELMo is better than BERT in sentence-level discourse parsing and \newcite{akbik-etal-2019-pooled} reported that LSTM with GloVe is better than BERT in named entity recognition.
As \newcite{clarke-lapata-2007-modelling} discussed, discourse and named entity information are both important in the sentence compression task.
Therefore, our observation is consistent with the previous researches.
These observations indicate that the best choice of word embedding types depends on a task.

Table \ref{tb:analysis} shows the actual outputs from each model.
In the first example, we can see that only \textbf{SLAHAN} can compress the sentence correctly.
However, both \textbf{Parent} and \textbf{Child} lack the words ``regulatory'' and ``full'', respectively, 
because \textbf{Parent} and \textbf{Child} can track only either parent or child words.
This result indicates that the selective gate module of \textbf{SLAHAN} can work well in a long sentence.

In the second example, \textbf{SLAHAN} and \textbf{Parent} compress the sentence correctly, whereas \textbf{Child} wrongly drops the words ``Broadway \textquotesingle s''.
This is because \textbf{Child} cannot explicitly track ``Jennifer Holliday'' from ``Broadway \textquotesingle s'' in the dependency tree.
This result also indicates that the selective gate of \textbf{SLAHAN} correctly switches the tracking direction from the parent or child in this case.

In the third example, only {\bf Child} can compress the sentence correctly.
This is because in this sentence the model can mostly retain important words by tracking only child words for each decoding step, as shown in Fig.\ref{fig:exp}.
In contrary, \textbf{SLAHAN}'s compressed sentence lacks the words ``US2 rescue''.
Because \textbf{SLAHAN} decides to use either the parent or child dependency graph by  using the selective gate module, we can understand that this wrong deletion is caused by the incorrect weights at the selective gate.
This result suggests that for compressing sentences more correctly, we need to make further improvement to the selective gate module.

\section{Related Work}

In the sentence compression task, many researches have adopted tree trimming methods \cite{jing:2000:ANLP,knight2000statistics,bergkirkpatrick-gillick-klein:2011:ACL-HLT2011,filippova-altun:2013:EMNLP}. 
As an alternative, LSTM-based models \cite{filippova-EtAl:2015:EMNLP,klerke-goldberg-sogaard:2016:N16-1} were introduced to avoid the effect of parsing errors in the tree trimming approach.
For using syntactic information in LSTM-based models, \newcite{filippova-EtAl:2015:EMNLP} additionally proposed a method to use parent words on a parsed dependency tree to compress a sentence.
\newcite{wang-EtAl:2017:Long4} used a LSTM-based tagger as a score function of an ILP-based tree trimming method to avoid the overfitting to the in-domain dataset.
These approaches have a merit to capture the syntactic information explicitly, but they were affected by parsing errors.

\newcite{kamigaito-etal-2018-higher} proposed a Seq2Seq model that can consider the higher-order dependency parents by tracking the dependency tree with their attention distributions.
Unlike the previous models, their model can avoid parse errors by jointly learning the summary generation probability and the dependency parent probability.
Similarly, \newcite{zhao-etal-2018-language} proposed a syntax-based language model that can compress sentences without using explicit parse trees.

Our SLAHAN uses strong language-model features, ELMo and BERT, and can track both parent and child words in a dependency tree without being affected by parse errors.
In addition, SLAHAN can retain important words by explicitly considering words that will be decoded in the future with our selective gate module during the decoding.

\section{Conclusion}
In this paper, we proposed a novel Seq2Seq model, \textit{syntactically look-ahead attention network} (SLAHAN), that can generate informative summaries by explicitly tracking parent and child words for capturing the important words in a sentence. 
The evaluation results showed that SLAHAN achieved the best  kept-token-based-F1, ROUGE-1, ROUGE-2 and ROUGE-L scores on the Google dataset in both all the sentence and the long sentence settings.
In the BNC corpus, SLAHAN also achieved the best  kept-token-based-F1, ROUGE-1, ROUGE-2 and ROUGE-L scores, and showed its effectiveness on both long sentences and out-of-domain sentences.
In human evaluation, SLAHAN improved informativeness without losing readability.
From these results, we can conclude that in Seq2Seq models, capturing important words that will be decoded in the future based on dependency relationships can help to compress long sentences during the decoding steps.

\section{Acknowledgement}
We are thankful to Dr. Tsutomu Hirao for his useful comments.

\fontsize{9.0pt}{10.0pt} \selectfont
\bibliographystyle{aaai}
\bibliography{aaai2020}

\appendix
\section*{Appendix}
\section*{A. Results of all sentences on the BNC Corpus}
We also report the results of all sentences on the BNC Corpus in Table \ref{tb:eval:bnc:all}.
\begin{table}[h]
\centering
\small
    \resizebox {\columnwidth} {!} {
\begin{tabular}{llccccc}
\toprule
 & & $\mathbf{F_1}$ & \thead{R-1} & \thead{R-2} & \thead{R-L} & $\mathbf{\Delta C}$ \\
\midrule
\multicolumn{2}{l}{Tagger}  & 68.4 & 56.7 & 44.4 & 56.5 & -24.6 \\
\multicolumn{2}{l}{LSTM}    & 67.4 & 54.8 & 42.7 & 54.6 & -27.0 \\
\multicolumn{2}{l}{LSTM-Dep}& 68.0 & 55.6 & 43.7 & 55.3 & -26.2 \\
\multicolumn{2}{l}{Attn}    & 67.1 & 54.3 & 43.1 & 54.1 & -26.5 \\
\multicolumn{2}{l}{Base}    & 68.3 & 56.0 & 43.9 & 55.8 & -25.6 \\
Parent & w/ syn             & 67.7 & 55.7 & 43.7 & 55.5 & -25.8 \\
Parent & w/o syn            & 67.5 & 55.2 & 43.1 & 55.0 & -26.5 \\
\midrule
Child & w/ syn              & 68.1 & 55.7 & 43.2 & 55.4 & -25.8 \\
Child & w/o syn             & 67.2 & 54.4 & 43.7 & 54.2 & -25.9 \\
SLAHAN & w/ syn             & $\mathbf{69.4}^{\dagger}$ & $\mathbf{57.6}^{\dagger}$ & $\mathbf{45.2}^{\dagger}$ & $\mathbf{57.3}^{\dagger}$ & $\mathbf{-23.7}^{\dagger}$ \\
SLAHAN & w/o syn            & 67.5 & 55.2 & 43.9 & 55.0 & -25.9 \\
\bottomrule
\end{tabular}}
    \caption{The bold values indicate the best scores. $\dagger$ indicates that the difference of the score from the best baseline is statistically significant. We used paired-bootstrap-resampling with 1,000,000 random samples for the significance test ($p<0.05$). \label{tb:eval:bnc:all}}
\end{table}

\newpage
\section*{B. Compression ratios in characters}

We used compression ratios in tokens to evaluate each method in this paper.
However, compression ratios in characters are also used for evaluating sentence compression performance.
Thus, we also report compression ratios in characters of our methods to support a fair comparison between sentence compression methods.
Table \ref{tb:eval:google:cr} and Table \ref{tb:eval:bnc:cr} show compression ratios in characters (\textbf{CR}) of methods for each setting in this paper.
Note that in both tables, $\mathbf{\Delta C}$ is calculated with compression ratios in characters.

\begin{table}[h]
\centering
\small
\begin{tabular}{llcccc}
\toprule
 & & \multicolumn{2}{c}{\textbf{ALL}} & \multicolumn{2}{c}{\textbf{LONG}} \\
 \cmidrule(lr){3-4}\cmidrule(lr){5-6}
 & & \textbf{CR} & $\mathbf{\Delta C}$ & \textbf{CR} & $\mathbf{\Delta C}$ \\
\midrule
\multicolumn{2}{l}{Gold}    & 42.3 & 0.0 & 30.9 & 0.0 \\
\midrule
\multicolumn{2}{l}{Tagger}  & 39.1 & -3.2 & 27.3 & -3.6 \\
\multicolumn{2}{l}{LSTM}    & 38.9 & -3.4 & 26.9 & -4.0 \\
\multicolumn{2}{l}{LSTM-Dep}& 38.9 & -3.4 & 27.1 & -3.8 \\
\multicolumn{2}{l}{Attn}    & 38.3 & -4.0 & 26.7 & -4.2 \\
\multicolumn{2}{l}{Base}    & 39.4 & -2.9 & 27.5 & -3.4 \\
Parent & w/ syn             & 39.7 & -2.6 & 27.3 & -3.6 \\
Parent & w/o syn            & 38.9 & -3.4 & 26.9 & -4.0 \\
\midrule
Child & w/ syn              & 39.3 & -3.0 & 27.3 & -3.6 \\
Child & w/o syn             & 39.1 & -3.2 & 27.5 & -3.4 \\
SLAHAN & w/ syn             & 40.7 & $\mathbf{-1.6}^{\dagger}$ & 28.4 & $\mathbf{-2.5}^{\dagger}$ \\
SLAHAN & w/o syn            & 39.1 & -3.2 & 27.3 & -3.6 \\
\bottomrule
\end{tabular}
    \caption{Compression ratios in characters on the Google dataset. The notations are the same as in Table \ref{tb:eval:bnc:all}.\label{tb:eval:google:cr}}
\end{table}
\begin{table}[h]
\centering
\small
\begin{tabular}{llcccc}
\toprule
 & & \multicolumn{2}{c}{\textbf{ALL}} & \multicolumn{2}{c}{\textbf{LONG}} \\
 \cmidrule(lr){3-4}\cmidrule(lr){5-6}
 & & \textbf{CR} & $\mathbf{\Delta C}$ & \textbf{CR} & $\mathbf{\Delta C}$ \\
\midrule
\multicolumn{2}{l}{Gold}    & 76.5 & 0.0 & 71.6 & 0.0 \\
\midrule
\multicolumn{2}{l}{Tagger}  & 51.2 & -25.3 & 31.2 & -40.4 \\
\multicolumn{2}{l}{LSTM}    & 48.2 & -28.3 & 31.0 & -40.6 \\
\multicolumn{2}{l}{LSTM-Dep}& 49.4 & -27.1 & 31.4 & -40.2 \\
\multicolumn{2}{l}{Attn}    & 48.5 & -28.0 & 30.5 & -41.1 \\
\multicolumn{2}{l}{Base}    & 49.9 & -26.6 & 31.8 & -39.8 \\
Parent & w/ syn             & 49.8 & -26.7 & 31.0 & -40.6 \\
Parent & w/o syn            & 49.0 & -27.5 & 30.1 & -41.5 \\
\midrule
Child & w/ syn              & 49.6 & -26.9 & 32.0 & -39.6 \\
Child & w/o syn             & 48.4 & -28.1 & 31.1 & -40.5 \\
SLAHAN & w/ syn             & 51.7 & $\mathbf{-24.8}^{\dagger}$ & 34.4 & $\mathbf{-37.2}^{\dagger}$ \\
SLAHAN & w/o syn            & 49.3 & -27.2 & 30.7 & -40.9 \\
\bottomrule
\end{tabular}
    \caption{Compression ratios in characters on the BNC Corpus. The notations are the same as in Table \ref{tb:eval:bnc:all}. \label{tb:eval:bnc:cr}}
\end{table}

\end{document}